\documentclass[runningheads]{llncs}
\usepackage{graphicx}
\usepackage{todonotes}
\usepackage{wrapfig}
\usepackage{caption}
\usepackage{subcaption}
\begin{document}
\title{Linked Credibility Reviews for Explainable Misinformation Detection}
%
%
\author{Ronald Denaux\inst{1}\orcidID{0000-0001-5672-9915} \and
 Jose Manuel Gomez-Perez\inst{1}\orcidID{0000-0002-5491-6431}}
\authorrunning{R. Denaux and JM Gomez-Perez}
%
\institute{Expert System, Madrid, Spain
\email{\{rdenaux,jmgomez\}@expertsystem.com}}
\maketitle              
\begin{abstract}
In recent years, misinformation on the Web has become increasingly rampant. The research community has responded by proposing systems and challenges, which are beginning to be useful for (various subtasks of) detecting misinformation. However, most proposed systems are based on deep learning techniques which are fine-tuned to specific domains, are difficult to interpret and produce results which are not machine readable. This limits their applicability and adoption as they can only be used by a select expert audience in very specific settings. In this paper we propose an architecture based on a core concept of Credibility Reviews (CRs) that can be used to build networks of distributed bots that collaborate for misinformation detection. The CRs serve as building blocks to compose graphs of (i) web content, (ii) existing credibility signals --fact-checked claims and reputation reviews of websites--, and (iii) automatically computed reviews. We implement this architecture on top of lightweight extensions to Schema.org and services providing generic NLP tasks for semantic similarity and stance detection.  Evaluations on existing datasets of social-media posts, fake news and political speeches demonstrates several advantages over existing systems: extensibility, domain-independence, composability, explainability and transparency via provenance. Furthermore, we obtain competitive results without requiring finetuning and establish a new state of the art on the Clef'18 CheckThat! Factuality task.
\end{abstract}
\section{Introduction}
\label{intro}

Although misinformation is not a new problem, the Web --due to the pace of news cycles combined with social media, and the information bubbles it creates-- has increasingly evolved into an ecosystem where misinformation can thrive\cite{marwick2017media} with various societal effects. Tackling misinformation\footnote{https://ec.europa.eu/digital-single-market/en/tackling-online-disinformation} is not something that can be achieved by a single organization --as evidenced by struggling efforts by the major social networks-- as it requires decentralisation, common conceptualisations, transparency and collaboration\cite{cazalens2018}.

Technical solutions for computer-aided misinformation detection and fact-checking have recently been proposed\cite{fullfact2016factchecking,hassan2017claimbuster} and are essential due to the scale of the Web. However, a lack of hand-curated data, maturity and scope of current AI systems, means assessing \emph{veracity}\cite{papadopoulos2016veracity} is not feasible. Hence the value of the current systems is not so much their accuracy, but rather their capacity of retrieving potentially relevant information that can help human fact-checkers, who are the main intended users of such systems, and are ultimately responsible for verifying/filtering the results such systems provide. Therefore, a main challenge is developing automated systems which can help the \emph{general public}, and influencers in particular, to assess the credibility of web content, which requires explainable results by AI systems. This points towards the need for hybrid approaches that enable the use of the best of deep learning-based approaches, but also of symbolic knowledge graphs to enable better collaboration between large platforms, fact-checkers, the general public and other stakeholders like policy-makers, journalists, webmasters and influencers.

In this paper, we propose a design on how to use semantic technologies to aid in resolving such challenges. Our contributions are:
\begin{itemize}
\item a datamodel and architecture of distributed agents for composable credibility reviews, including a lightweight extension to \texttt{schema.org} to support provenance and explainability (section~\ref{LCR})
\item an implementation of the architecture demonstrating feasibility and value (section~\ref{acred})
\item an evaluation on various datasets establishing state-of-the-art in one dataset (Clef'18 CheckThat! Factuality task) and demonstrating capabilities and limitations of our approach, as well as paths for improvements (section~\ref{eval})
\end{itemize}

\section{Related Work}
\label{related-work}

The idea of automating (part of) the fact-checking process is relatively recent\cite{fullfact2016factchecking}. ClaimBuster\cite{hassan2017claimbuster} proposed the first automated fact-checking system and its architecture is mostly still valid, with a database of fact-checks and components for monitoring web sources, spotting claims and matching them to previously fact-checked claims. Other similar services and projects include Truly media\footnote{\url{https://www.truly.media/} \url{https://www.disinfobservatory.org/}}, invid\footnote{\url{https://invid.weblyzard.com/}} and CrowdTangle\footnote{\url{https://status.crowdtangle.com/}}.  These systems are mainly intended to be used by professional fact-checkers or journalists, who can evaluate whether the retrieved fact-check article is relevant for the identified claim. These automated systems rarely aim to predict the accuracy of the content; this is (rightly) the job of the journalist or fact-checker who uses the system. Many of these systems provide valuable REST APIs to access their services, but as they use custom schemas, they are difficult to compose and inspect as they are not machine-interpretable or explainable.

Besides full-fledged systems for aiding in fact-checking, there are also various strands of research focusing on specific computational tasks needed to identify misinformation or assess the accuracy or veracity of web content based on ground credibility signals. Some low-level NLP tasks include check-worthiness\cite{nakov2018clefCheckThat} and stance detection\cite{Schiller2020,pomerleau2017fakeNewsChallenge}, while others aim to use text classification as a means of detecting deceptive language\cite{perez2018fakenews}. Other tasks mix linguistic and social media analysis, for example to detect and classify rumours\cite{zubiaga2018rumours}. Yet others try to assess veracity of a claim or document by finding supporting evidence in (semi)structured data\cite{thorne-etal-2019-fever2}. These systems, and many more, claim to provide important information needed to detect misinformation online, often in some very specific cases. However without a clear conceptual and technical framework to integrate them, the signals such systems provide are likely to go unused and stay out of reach of users who are exposed to misinformation.

The Semantic Web and Linked Data community has also started to contribute ideas and technical solutions to help in this area: perhaps the biggest impact has been the inclusion in Schema.org\cite{guha2016schema} of the \texttt{ClaimReview} markup\footnote{\url{https://www.blog.google/products/search/fact-check-now-available-google-search-and-news-around-world/}}, which enables fact-checkers to publish their work as machine readable structured data. This has enabled aggregation of such data into knowledge graphs like ClaimsKG~\cite{Tchechmedjiev2019}, which also performs much needed normalisation of labels, since each fact-checker uses its own set of labels.
A conceptual model and RDF vocabulary was proposed to distinguish between the utterance and propositional aspects of claims~\cite{boland2019modelingClaims}. It allows expressing fine-grained provenance (mainly of annotations on the text of the claim), but still relies on \texttt{ClaimReview} as the main accuracy describing mechanism. It is unclear whether systems are actually using this RDF model to annotate and represent claims as the model is heavyweight and does not seem to align well with mainstream development practices. In this paper, we build on these ideas to propose a lightweight model which focuses on the introduction of a new type of Schema.org Review that focuses on \emph{credibility} rather than \emph{factuality}\footnote{In our opinion, current AI systems cannot truly assess veracity since this requires human skills to access and interpret new information and relate them to the world.}.

The focus on credibility, defined as an estimation of factuality based on available signals or evidence, is inspired by MisinfoMe\cite{mensio2019NewsSourceCredibility,mensio2019misinfome} which borrows from social science, media literacy and journalism research. MisinfoMe focuses on credibility of sources, while we expand this to credibility of any web content and integrate some of MisinfoMe's services in our implementation to demonstrate how our approach enables composition of such services. There is also ongoing work on W3C Credibilty
Signals\footnote{\url{https://credweb.org/signals-beta/}}, which aims to define a vocabulary to specify \emph{credibility indicators} that may be relevant for assessing the credibility of some web content. To the best of our knowledge, this is still work in progress and no systems are implementing the proposed vocabularies.

\section{Linked Credibility Reviews}
\label{LCR}

This section presents \emph{Linked Credibility Reviews} (LCR), our proposed linked data model for composable and explainable misinformation detection. As the name implies, \emph{Credibility Review}s (CR) are the main resources and outputs of this architecture. We define a CR as a tuple $\langle d, r, c, p \rangle$, where the CR:
\begin{itemize}
  \item reviews a \emph{data item} $d$, this can be any linked-data node but will typically refer to articles, claims, websites, images, social media posts, social media accounts, people, publishers, etc.
  \item assigns a \emph{credibility rating} $r$ to the \emph{data item} under review and qualifies that rating with a \emph{rating confidence} $c$. 
  \item \emph{provides provenance information} $p$ about:
    \begin{itemize}
      \item \emph{credibility signals} used to derive the credibility rating. Credibility Signals (CS) can be either (i) CRs for data items relevant to the data item under review or (ii) \emph{ground credibility signals} (GCS) resources (which are not CRs) in databases curated by a trusted person or organization.
      \item the \emph{author} of the review. The author can be a person, organization or bot. Bots are automated agents that produce CRs for supported data items based on a variety of strategies, discussed below.
    \end{itemize}
\end{itemize}

The credibility rating is meant to provide a subjective (from the point-of-view of the author) measure of how much the credibility signals support or refute the content in data item. Provenance information is therefore crucial as it allows humans ---e.g. end-users, bot developers--- to retrace the CRs back to the ground credibility signals and assess the accuracy of the (possibly long) chain of bots (and ultimately humans) that were involved in reviewing the initial data item. It also enables the generation of explanations for each step of the credibility review chain in a composable manner as each bot (or author) can describe its own strategy to derive the credibility rating based on the used credibility signals.

\paragraph{Bot Reviewing Strategies}
CR bots are developed to be able to produce CRs for specific data item types. We have identified a couple of generic strategies that existing services seem to implement and which can be defined in terms of CRs (these are not exhaustive, though see figure~\ref{fig:pipeline} for a depiction of how they can collaborate):
\begin{itemize}
\item \textbf{ground credibility signal lookup} from some trusted source. CR bots that implement this strategy will (i) generate a query based on $d$ and (ii) convert the retrieved ground credibility signal into a CR;
\item \textbf{linking} the item-to-review $d$ with $n$ other data items $d'_i$ \emph{of the same type}, for which a $\mathrm{CR}_{d'_i}$ is available. These bots define functions $f_r$, $f_c$ and $f_\mathrm{agg}$. The first two, compute the new values $r_i$ and $c_i$ based on the original values and the relation or similarity between $d$ and $d'_i$ i.e. $r_i = f_r(\mathrm{CR}_{d'_i}, d, d')$. These produce $n$ credibility reviews, $\mathrm{CR}^i_d$, which are then aggregated into $\mathrm{CR}_{d} = f_{\mathrm{agg}}(\{CR^i_d \; | \; 0 \le i < n\})$.
\item \textbf{decomposing} whereby the bot identifies relevant parts $d'_i$ of the item-to-review $d$ and requests CRs for those parts $CR_{d'_i}$. Like the linking bots, these require deriving new credibility ratings $CR_{d_i}$ and confidences based on the relation between the whole and the parts; and aggregating these into the CR for the whole item. The main difference is that the parts can be items of different types.
\end{itemize}

\paragraph{Representing and Aggregating Ratings}
For ease of computation, we opt to represent credibility ratings and their confidences as follows:
\begin{itemize}
\item $r \in \Re$, must be in the range of $[-1.0, 1.0]$ where $-1.0$ means not credible and $1.0$ means credible
\item $c \in \Re$, must be in the range of $[0.0, 1.0]$ where $0.0$ means no confidence at all and $1.0$ means full confidence in the accuracy of $r$, based on the available evidence in $p$.
\end{itemize}

This representations makes it possible to define generic, relatively straightforward aggregation functions like:
\begin{itemize}
  \item $f^\mathrm{mostConfident}$ which selects $\mathrm{CR}_i$ which has the highest confidence value $c$
  \item $f^\mathrm{leastCredible}$ which selects the $\mathrm{CR}_i$ which has the lowest value $r$
\end{itemize}

\subsection{Extending \texttt{schema.org} for LCR}
While reviewing existing ontologies and vocabularies which could be reused to describe the LCR model, we noticed that \texttt{schema.org}\cite{guha2016schema} was an excellent starting point since it already provides suitable schema types for data items on the web for which credibility reviews would be beneficial (essentially any schema type that extends \texttt{CreativeWork}). It already provides suitable types for \texttt{Review}, \texttt{Rating}, as well as properties for expressing basic provenance information and meronymy (\texttt{hasPart}). Some basic uses and extensions compliant with the original definitions are:

\begin{itemize}
  \item Define \texttt{CredibilityReview} as an extension of \texttt{schema:Review}, whereby the \texttt{schema:reviewAspect} is \texttt{credibility}\footnote{Note that \texttt{ClaimReview} is not suitable since it is overly restrictive: it can only review \texttt{Claim}s (and it assumes the review aspect is, implicitly, \texttt{accuracy}).}
  \item use \texttt{schema:ratingValue} to encode $r$
  \item add a \texttt{confidence} property to \texttt{schema:Rating} which encodes the rating confidence $c$.
  \item use \texttt{isBasedOn} to record that a CR was computed based on other CRs. We also use this property to describe dependencies between CR Bots, even when those dependencies have not been used as part of a CR.
  \item use \texttt{author} to link the CR with the bot that produced it
\end{itemize}

The main limitation we encountered with the existing definitions was that \texttt{CreativeWork}s (including \texttt{Review}s) are expected to be created only by \texttt{Person}s or \texttt{Organization}s, which excludes reviews created by bots. We therefore propose to extend this definition by:
\begin{itemize}
  \item introducing a type \texttt{Bot} which extends \texttt{SoftwareApplication}
  \item allowing \texttt{Bot}s to be the \texttt{author}s of \texttt{CreativeWork}s.
\end{itemize}

\begin{figure}
  \begin{center}
     \includegraphics[width=0.8\textwidth]{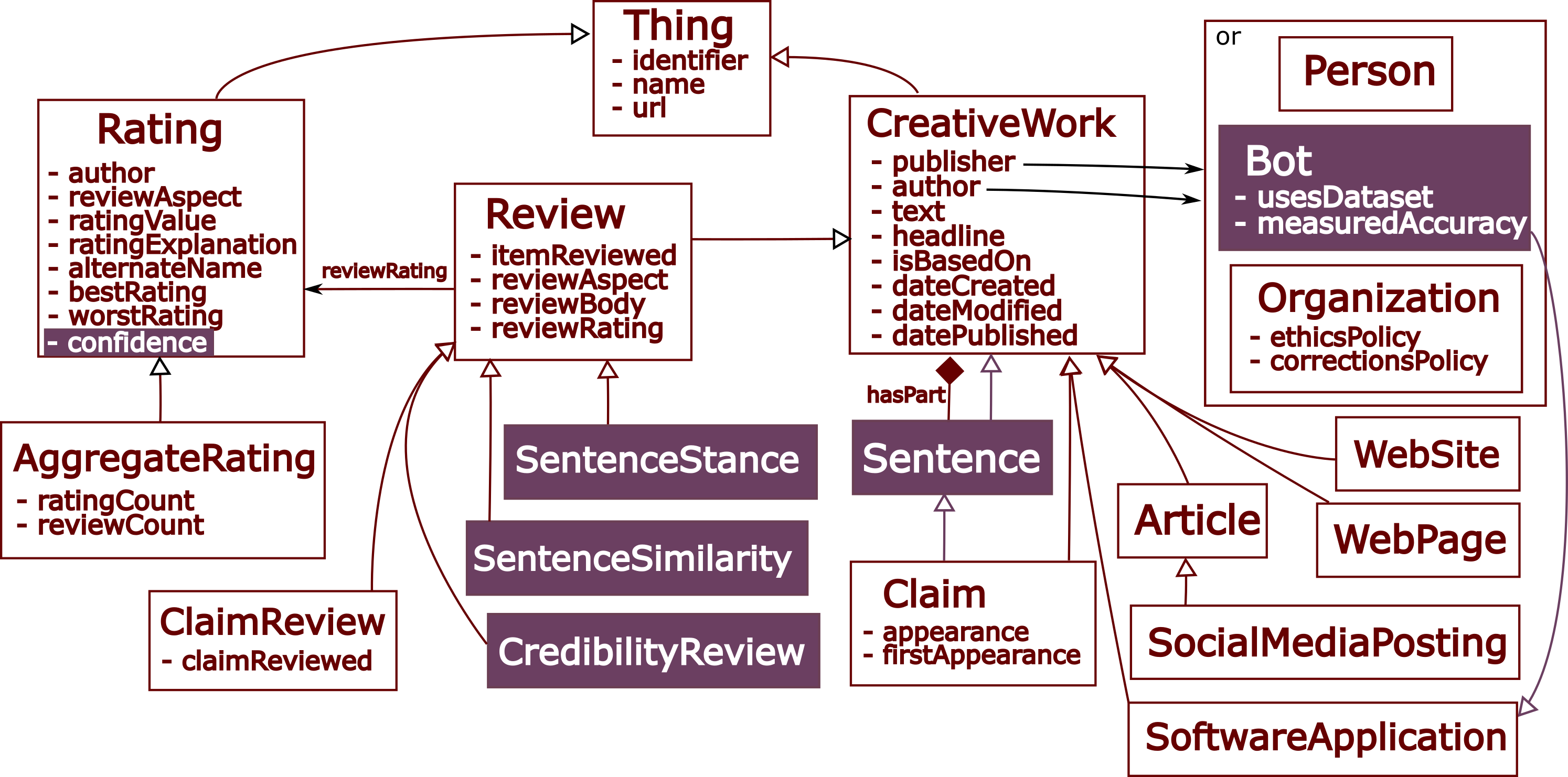}
  \end{center}
  \caption{Linked Credibility Review data model, extending schema.org.}
  \label{fig:data-model}
\end{figure}

Finally, in this paper we focus on \emph{textual} misinformation detection and found we were missing a crucial type of \texttt{CreativeWork}, namely \texttt{Sentence}s. Recently, a \texttt{Claim} type was proposed, to represent factually-oriented sentences and to work in tandem with the existing \texttt{ClaimReview}, however, automated systems still have trouble determining whether a sentence is a claim or not, therefore, CR bots should be able to review the credibility of \texttt{Sentence}s and relevant aspects between pairs of sentences such as their stance and similarity. The overall \texttt{schema.org} based data model is depicted in figure~\ref{fig:data-model}, focused on CRs for textual web content (we leave other modalities as future work).

\section{acred -- Deep Learning-based CR bots}
\label{acred}

To demonstrate the potential of the Linked Credibility Review architecture, we have implemented a series of CR bots capable of collaborating to review articles, tweets, sentences and websites.\footnote{The source code is available at \url{https://github.com/rdenaux/acred}} We present the conceptual implementation in sections \ref{sec:acred-gcs} to
\ref{sec:acred-decbots} and provide further details in section~\ref{sec:implDetails}.

\subsection{Ground Credibility Signal Sources}
\label{sec:acred-gcs}
Ultimately we rely on two ground credibility signal sources:
\begin{itemize}
\item A database of \texttt{ClaimReview}s which provide accuracy ratings for factual claims by a variety of fact-checkers. 
\item Third-party, well-established services for validating \texttt{WebSite}s, such as NewsGuard and Web Of Trust\footnote{\url{https://www.newsguardtech.com/}, \url{https://www.mywot.com/}}, which rely on either expert or community-based ratings.
\end{itemize}

\subsection{GCS Lookup Bots}
The two GCS sources are mediated via two GCS lookup bots.

The $\mathtt{LookupBot}_{\mathtt{ClaimReview}}$ returns a CR for a \texttt{Claim} based on a \texttt{ClaimReview} from the database. In order to derive a CR from a \texttt{ClaimReview}, the accuracy rating in the \texttt{ClaimReview} need to be converted into equivalent credibility ratings. The challenge here is that each fact-checker can encode their review rating as they see fit. The final review is typically encoded as a textual \texttt{alternateName}, but sometimes also as a numerical \texttt{ratingValue}. ClaimsKG already performs this type of normalisation into a set of ``veracity'' labels, but for other \texttt{ClaimReview}s we have developed a list of simple heuristic rules to assign a credibility and confidence score.

The $\mathtt{LookupBot}_\mathtt{WebSite}$ returns a CR for a \texttt{WebSite}. This is a simple wrapper around the existing MisinfoMe aggregation service~\cite{mensio2019NewsSourceCredibility}, which already produces credibility and confidence values.

\subsection{Linking Bots}
Although the GCS lookup bots provide access to the basic credibility signals, they can only provide this for a relatively small set of claims and websites. Misinformation online often appears as  variations of fact-checked claims and can appear on a wide variety of websites that may not have been reviewed yet by a human. Therefore, to increase the number of sentences which can be reviewed, we developed the following linking bots (see Section~\ref{sec:implDetails} for further details).

The $\mathtt{LinkBot}^{\mathtt{PreCrawled}}_{\mathtt{Sentence}}$ uses a database of pre-crawled sentences extracted from a variety of websites.  A proprietary NLP system\footnote{\url{http://expert.ai}} extracts the most relevant sentences that may contain factual information in the crawled documents. This is done by identifying sentences that (i) are associated with a topic (e.g. Politics or Health) and (ii) mentions an entity (e.g. a place or person). The CR for the extracted sentence is assigned based on the website where the sentence was found (i.e. by using the $\mathtt{LookupBot}_\mathtt{WebSite}$). Since not all sentences published by a website are as credible as the site, the resulting CR for the sentence has a lower confidence than the CR for the website itself.

The $\mathtt{LinkBot}^{\mathtt{SemSim}}_\mathtt{Sentence}$ is able to generate CRs for a wide variety of sentences by linking the input sentence $s_i$ to sentences $s_j$ for which a CR is available (via  other bots). This linking is achieved by using a neural sentence encoder $f_{\mathrm{enc}}: S \mapsto \Re^d$ --where $S$ is the set of sentences and $d \in N^+$--, that is optimised to encode semantically similar sentences close to each other in an embedding space. The bot creates an index by generating embeddings for all the sentences reviewed by the $\mathtt{LookupBot}_{\mathtt{ClaimReview}}$ and the $\mathtt{LinkBot}^{\mathtt{PreCrawled}}_{\mathtt{Sentence}}$. The incoming sentence, $s_i$ is encoded and a nearest neighbor search produces the closest matches along with a similarity score based on a similarity function $f_\mathrm{sim}: \Re^d \times \Re^d \mapsto \Re^{[0,1]}$. Unfortunately, most sentence encoders are not polarity aware so that negations of a phrase are considered similar to the original phrase; therefore we use a second neural model for stance detection $f_\mathrm{stance}: S \times S \mapsto \mathrm{SL}$, where $\mathrm{SL}$ is a set of stance labels. We then define $f_\mathrm{polarity}: \mathrm{SL} \mapsto \{1, -1\}$, which we use to invert the polarity of $r_j$ if $s_i$ disagrees with $s_j$. We also use the predicted stance to revise the similarity score between $s_i$ and $s_j$ by defining a function $f_\mathrm{reviseSim} : \mathrm{SL}, \Re^{[0,1]} \mapsto \Re^{[0,1]}$. For example, stances like \emph{unrelated} or \emph{discuss} may reduce the estimated similarity, which is used to revise the confidence of the original credibility. In summary, the final CR for $s_i$ is selected via $f^\mathrm{mostConfident}$ from a pool of $\mathrm{CR}_{i,j}$ for the matching $s_j$; where the rating and confidences for each $\mathrm{CR}_{i,j}$ are given by:
\[ r_i = r_j \times f_\mathrm{polarity}(f_\mathrm{stance}(s_i, s_j)) \]
\[ c_i = c_j \times f_\mathrm{reviseSim}\Big(f_\mathrm{stance}(s_i, s_j), f_\mathrm{sim}\big(f_\mathrm{enc}(s_i), f_\mathrm{enc}(s_j)\big)\Big)\]

\subsection{Decomposing Bots}
\label{sec:acred-decbots}
By combining linking and GCS lookup bots we are already capable of reviewing a wide variety of sentences. However, users online encounter misinformation in the form of high-level \texttt{CreativeWork}s like social media posts, articles, images, podcasts, etc. Therefore we need bots which are capable of (i) dissecting those CreativeWorks into relevant parts for which CRs can be calculated and (ii) aggregating the CRs for individual parts into an overall CR for the whole \texttt{CreativeWork}. In acred, we have defined two main types:
\begin{itemize}
\item $\mathtt{DecBot}_\mathtt{Article}$ reviews \texttt{Article}s, and other long-form textual \texttt{CreativeWork}s
\item $\mathtt{DecBot}_\mathtt{SocMedia}$ reviews \texttt{SocialMediaPost}s
\end{itemize}

In both bots, decomposition works by performing  NLP and content analysis on the title and textual content of the \texttt{CreativeWork} $d_i$. This results in a set of parts $P = \{d_j\}$ which include \texttt{Sentence}s, linked \texttt{Article}s or \texttt{SocialMediaPost}s and metadata like the \texttt{WebSite} where $d$ was published. Each of these can be analysed either recursively or via other bots, which results in a set of reviews $\{\mathrm{CR}_j\}$ for the identified parts. We define a function $f_{\mathrm{part}}$ which maps $\mathrm{CR}_j$ onto $\mathrm{CR_{i,j}}$, which takes into account the relation between $d_i$ and $d_j$ as well as the provenance of $\mathrm{CR}_j$ to derive the credibility rating and confidence. The final $\mathrm{CR}_i$ is selected from all $\mathrm{CR}_{i,j}$ via $f^{\mathrm{leastCredible}}$.

\begin{figure}
  \begin{center}
    \includegraphics[width=\textwidth]{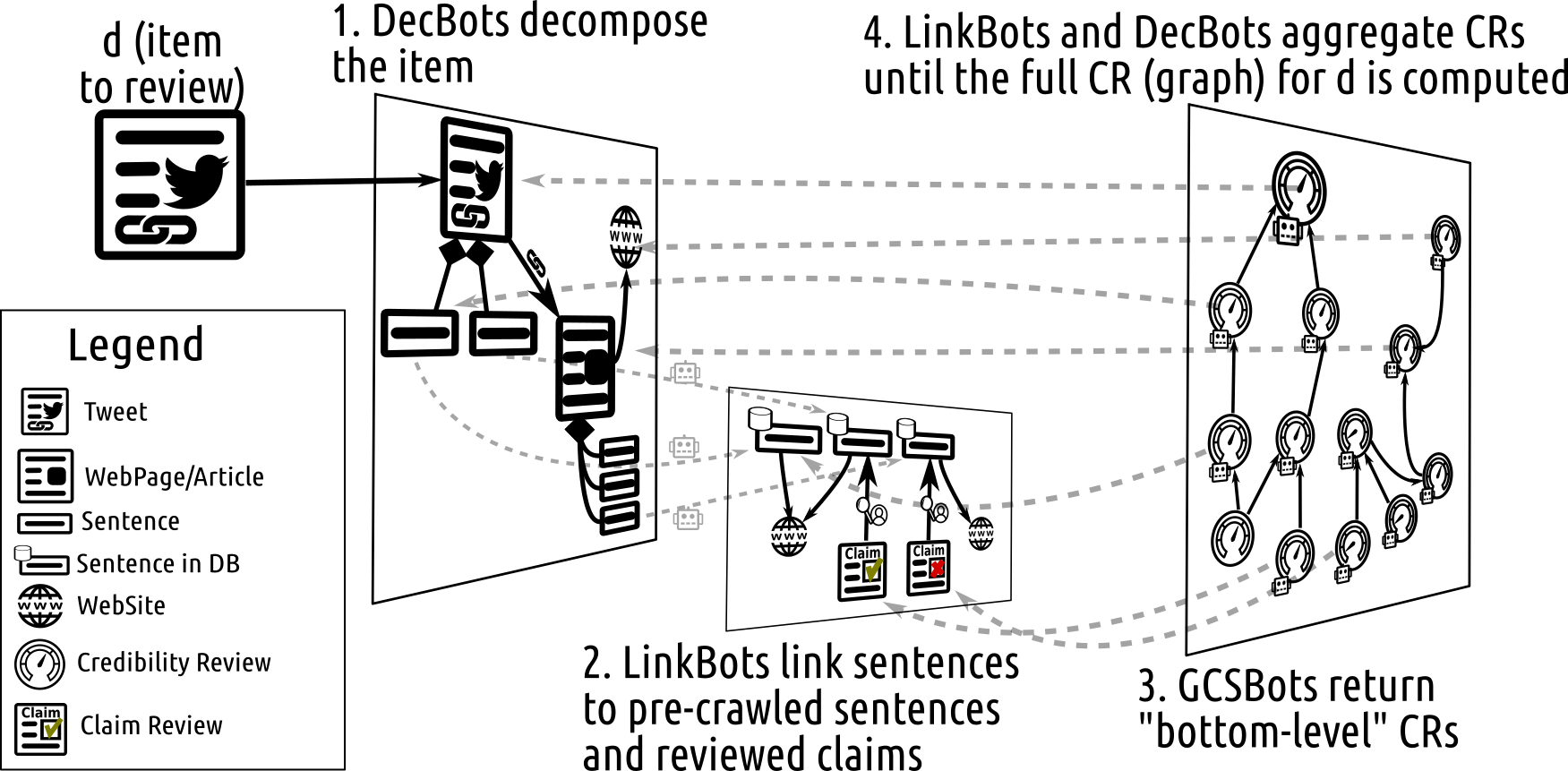}
  \end{center}
  \caption{Depiction of acred bots collaborating to produce a CR for a tweet.}
  \label{fig:pipeline}
\end{figure}

Figure~\ref{fig:pipeline} shows a diagram depicting how the various CR bots compose and collaborate to review a tweet.

\subsection{Implementation details}
\label{sec:implDetails}
 Our database of \texttt{ClaimReview}s contains 45K claims and was based on public resources such as ClaimsKG~\cite{Tchechmedjiev2019} (32K), data-commons (9.6K) and our in-house developed crawlers (4K). The database of pre-crawled sentences contained 40K sentences extracted from a variety of generic news sites on-line between april 2019 and april 2020. It consisted primarily in relatively well-regarded news sites like \texttt{expressen.se}, \texttt{krone.at}, \texttt{zdf.de}, which combined for about 35K of the sentences, and a long tail of other sites including \texttt{theconversation.com} and \texttt{heartland.org}. For reproducibility, we will publish the list of sentences along associated URLs.

Our heuristic rules to normalise \texttt{ClaimReviews} are implemented in about 50 lines of python to map numeric \texttt{ratingValue}s (in context of specified \texttt{bestRating} and \texttt{worstRating} values) and 150 lines of python to map about 100 \texttt{alternateName} values (e.g ``inaccurate'', ``false and misleading'', ``this is exaggerated'') and about 20 patterns (e.g. ``wrong.*'', ``no, *'') into estimated $c$, $r$ values.
 
The sentence encoder, $f_{\mathrm{enc}}$ in $\mathtt{LinkBot}^{\mathtt{SemSim}}_\mathtt{Sentence}$ is implemented as a RoBERTa-base~\cite{Liu2019RoBERTa} model finetuned on STS-B~\cite{Cer2017STS-B}. We employ a siamese structure as this enables us to perform encoding of the claims off-line (slow) and comparison on-line (fast) at the cost of some accuracy. Our model achieves $83\%$ Pearson correlation on STS-B dev.

The stance detector, $f_\mathrm{stance}$, is also a RoBERTa-base model trained on FNC-1~\cite{pomerleau2017fakeNewsChallenge}, which assigns a stance label (either ``agree'', ``disagree'', ``discuss'' or ``unrelated'') to pairs of texts: about 50K for training and about 25K for testing. Our model achieves $92\%$ accuracy on the held-out test set. See our GitHub repository for links to model weights and jupyter notebooks replicating our fine-tuning procedure.

 Each implemented bot defines a set of templates to generate textual explanations. These reflect processing performed by the bot in a way that can be inspected by a user. Produced CRs use the \texttt{schema:ratingExplanation} property to encode the generated explanations and use markdown to take advantage of hypertext capabilities like linking and formatting of the explanations. Examples are presented in table~\ref{tab.explanations}.

\begin{table}
\caption{Example explanations generated by our bots.}\label{tab.explanations}
\begin{tabular}{lp{10cm}}
Bot &  Example explanation \\
\hline
  $\mathtt{LookupBot}_{\mathtt{ClaimRev}}$ & \footnotesize{Claim \texttt{'Ford is moving all of their small-car productin to Mexico.'} is \emph{mostly not credible} based on a \href{http://www.politifact.com/truth-o-meter/statements/2016/oct/23/donald-trump/donald-trump-says-ford-moving-all-small-car-produc/}{fact-check} by \href{http://www.politifact.com}{politifact} with normalised numeric ratingValue 2 in range [1-5] } \\\noalign{\smallskip}
  
  $\mathtt{LookupBot}_\mathtt{WebSite}$ &  \footnotesize{Site \texttt{www.krone.at} seems \emph{mostly credible} based on 2 review(s) by external rater(s) \href{https://www.newsguardtech.com/}{NewsGuard} or \href{https://mywot.com/}{Web Of Trust}} \\\noalign{\smallskip}
  
  $\mathtt{LinkBot}^{\mathtt{PreCrawled}}_{\mathtt{Sentence}}$ & \footnotesize{Sentence \texttt{Now we want to invest in the greatest welfare program in modern times.} seems \emph{credible} as it was published in site \texttt{www.expressen.se}. (Explanation for WebSite omitted) }\\\noalign{\smallskip}
  
  $\mathtt{LinkBot}^{\mathtt{SemSim}}_\mathtt{Sentence}$ & \footnotesize{Sentence \texttt{When Senator Clinton or President Clinton asserts that I said that the Republicans had had better economic policies since 1980, that is not the case.} seems \emph{not credible} as it agrees with sentence:
                                                           
                                                           \texttt{Obama said that 'since 1992, the Republicans have had all the good ideas...'} that seems \emph{not credible} based on a \href{http://www.politifact.com/truth-o-meter/statements/2008/jan/21/bill-clinton/obama-not-a-reagan-democrat/}{fact-check} by \href{http://www.politifact.com}{politifact} with textual rating 'false'. Take into account that the sentence appeared in site \texttt{www.cnn.com} that seems \emph{credible} based on 2 review(s) by external rater(s) \href{https://www.newsguardtech.com/}{NewsGuard} or \href{https://mywot.com/}{Web Of Trust}}\\\noalign{\smallskip}

  $\mathtt{LinkBot}^{\mathtt{SemSim}}_\mathtt{Sentence}$ & \footnotesize{Sentence \texttt{Can we reduce our dependence on foreign oil and by how much in the first term, in four years?} is similar to and discussed by:

                                                           \texttt{Drilling for oil on the Outer Continental Shelf and in parts of Alaska will  'immediately reduce our dangerous dependence on foreign oil.'} that seems \emph{not credible}, based on a \href{http://www.politifact.com/ohio/statements/2011/may/04/rob-portman/sen-rob-portman-says-easing-access-drilling-would-/}{fact-check} by \href{http://www.politifact.com}{politifact} with textual rating 'false'. } \\\noalign{\smallskip}

  $\mathtt{DecBot}_\mathtt{Article}$ & \footnotesize{Article \href{http://www.cnn.com/2008/POLITICS/01/21/debate.transcript/index.html}{``Part 1 of CNN Democratic presidential debate''} seems \emph{not credible} based on its least credible sentence. (explanation for sentence CR omitted)} \\\noalign{\smallskip}
  $\mathtt{DecBot}_\mathtt{SocMedia}$ & \footnotesize{Sentence \texttt{Absolutely fantastic, there is know difference between the two facist socialist powers of today’s EU in Brussels, and the yesteryears of Nazi Germany} in \href{https://twitter.com/yhwhuniversity/status/1131891504522375168}{tweet} agrees with:

                \texttt{'You see the Nazi platform from the early 1930s ... look at it compared to the (Democratic Party) platform of today, you're saying, 'Man, those things are awfully similar.''} that seems \emph{not credible} based on a \href{http://www.politifact.com/truth-o-meter/statements/2018/aug/03/donald-trump-jr/did-nazi-platform-echo-democratic-platform-donald-/}{fact-check} by \href{http://www.politifact.com}{politifact} with textual claim-review rating 'false'"}\\
\hline
\end{tabular}
\end{table}

 Different CR bots are deployed as separate Docker images and expose a REST API accepting and returning JSON-LD formatted requests and responses. They are all deployed on a single server (64GB RAM, Intel i7-8700K CPU @ 3.70GHzn with 12 cores) via docker-compose. The \texttt{ClaimReview} and pre-crawled sentence databases are stored in a Solr instance. The index of encoded sentences is generated off-line on a separate server with a GPU by iterating over the claims and sentences in Solr, and loaded into memory on the main server at run-time.
 
\section{Evaluation}
\label{eval}
One of the main characteristics of the LCR architecture is that CR bots can be distributed across different organizations. This has the main drawback that it can be more difficult to fine-tune bots to specific domains since top-level bots do not have direct control on how lower-level bots are implemented and fine-tuned. Therefore in this paper we first evaluated our \texttt{acred} implementation, described above, on a variety of datasets covering social media posts, news articles and political speeches. Our rationale is that existing, non-distributed fact-checking approaches have an edge here as they can fine-tune their systems based on training data and therefore provide strong baselines to compare against. We used the explanations, along with the provenance trace, to perform error analysis\footnote{Note that usability evaluation of the generated explanations is not in the scope of this paper.} on the largest dataset, described in Section~\ref{sec:eval-results}. This showed \texttt{acred} was overly confident in some cases. To address this, we introduced a modified version, $\mathtt{acred}^\mathtt{+}$ with custom functions to reduce the confidence and rating values of two bots under certain conditions: $\mathtt{DecBot}_\mathtt{Article}$ when based only on a website credibility; $\mathtt{LinkBot}^{\mathtt{SemSim}}_\mathtt{Sentence}$ when the stance is ``unrelated'' or ``discuss''.

\subsection{Datasets}
The first dataset we use is the Clef'18 CheckThat! Factuality Task~\cite{nakov2018clefCheckThat} (\texttt{clef18}). The task consists in predicting whether a check-worthy claim is either \texttt{true}, \texttt{half-true} or \texttt{false}. The dataset was derived from fact-checked political debates and speeches by \texttt{factcheck.org}. For our evaluation we only use the English part of this dataset\footnote{Our implementation has support for machine translation of sentences, however this adds a confounding factor hence we leave this as future work.} which contains 74 and 139 claims for training and testing.

FakeNewsNet~\cite{shu2018FakeNewsNet} aims to provide a dataset of fake and real news enriched with social media posts and context sharing those news. In this paper we only use the fragment derived from articles fact-checked by Politifact that have textual content, which consists of 420 \emph{fake} and 528 \emph{real} articles. The articles were retrieved by following the instructions on the Github page\footnote{\url{https://github.com/KaiDMML/FakeNewsNet}, although we note that text for many of the articles could no longer be retrieved, making a fair comparison difficult.}.

Finally, \texttt{coinform250}\footnote{\url{https://github.com/co-inform/Datasets}} is a dataset of 250 annotated tweets. The tweets and original labels were first collected by parsing and normalising \texttt{ClaimReview}s from datacommons and scraping fact-checker sites using the MisinfoMe data collector~\cite{mensio2019misinfome,mensio2019NewsSourceCredibility}. Note that acred's collection system is \textbf{not} based on MisinfoMe\footnote{acred's data collector is used to build the \texttt{ClaimReview} database described in Sect.~\ref{acred}; it does not store the \texttt{itemReviewed} URL values; only the \texttt{claimReviewed} strings.}. The original fact-checker labels were mapped onto six labels (see table~\ref{tab:coinformLabels}) by 7 human raters achieving a Fleiss $\kappa$ score of $0.52$ (moderate agreement). The fine-grained labels make this a challenging but realistic dataset.

\begin{wraptable}{r}{0.6\linewidth}
  \caption{Mapping of credibility \texttt{ratingValue} $r$ and \texttt{confidence} $c$ for coinform250.}
  \label{tab:coinformLabels}
  \centering
  \begin{tabular}{lll}
    label & $r$ & $c$ \\
    \hline
    credible & $r \ge 0.5$ & $c > 0.7$ \\
    mostly credible & $0.5 > r \ge 0.25$ & $c > 0.7$ \\
    uncertain & $0.25 > r \ge -0.25$ & $c > 0.7$ \\
    mostly not credible & $-0.25 > r \ge -0.5$ & $c > 0.7$\\
    not credible & $-0.5 > r$ & $c > 0.7$ \\
    not verifiable & any & $c \le 0.7$ \\
    \hline
  \end{tabular}
\end{wraptable}

For each dataset our prediction procedure consisted in steps to (i) read samples, (ii) convert them to the appropriate \texttt{schema.org} data items (\texttt{Sentence}, \texttt{Article} or \texttt{SocialMediaPost}), (iii) request a review from the approriate acred CR bot via its REST API; (iv) map the produced CR onto the dataset labels  and (v) optionally store the generated graph of CRs. For \texttt{clef18} we set $t = 0.75$, so that $r >= t$ has label \texttt{TRUE}, $r <= -0.75$ has label \texttt{FALSE} and anything in between is \texttt{HALF-TRUE}. See table~\ref{tab:coinformLabels} for \texttt{coinform250} threshold definitions.

\subsection{Results}
\label{sec:eval-results}
On \texttt{clef18}, \texttt{acred} establishes a new state-of-the-art result as shown in table~\ref{tab:clef18Results}, achieving $0.6835$ in MAE, the official metric in the competition~\cite{nakov2018clefCheckThat}. This result is noteworthy as, unlike the other systems, \texttt{acred} did not use the training set of \texttt{clef18} at all to finetune the underlying models. With $\mathtt{acred}^\mathtt{+}$, we further improved our results achieving $0.6475$ MAE.

\begin{table}
  \caption{Results on \texttt{clef18} English test dataset compared to baselines. The bottom rows shows results on the English training set.}
  \label{tab:clef18Results}
  \centering
  \begin{tabular}{lrrrrr}
    system & MAE & Macro MAE & Acc & Macro F1 & Macro AvgR\\
    \hline
    \texttt{acred} & 0.6835 & 0.6990 & \textbf{0.4676} & \textbf{0.4247} & 0.4367\\
    $\mathtt{acred}^\mathtt{+}$ & \textbf{0.6475} & \textbf{0.6052} & 0.3813 & 0.3741 & 0.4202\\
    Copenhagen\cite{wang2018copenhagen} & 0.7050 & 0.6746 & 0.4317 & 0.4008 & \textbf{0.4502}\\
    random\cite{nakov2018clefCheckThat} & 0.8345 & 0.8139 & 0.3597 & 0.3569 & 0.3589\\
    \hline
    \texttt{acred} ``training'' & 0.6341 & 0.7092 & 0.4878 & 0.4254 & 0.4286\\
    $\mathtt{acred}^\mathtt{+}$ ``training'' & 0.6585 & 0.6386 & 0.4024 & 0.3943 & 0.4020\\
    \hline
  \end{tabular}
\end{table}


\begin{wraptable}{r}{0.6\linewidth}
  \caption{Results on FakeNewsNet Politifact compared to baselines that only use article content.}
  \label{tab:fakeNewsNetResults}
  \centering
  \begin{tabular}{lrrrr}
    System & Accuracy & Precision & Recall & F1 \\
    \hline
    \texttt{acred} & 0.586 & 0.499 & \textbf{0.823} & 0.622 \\
    $\mathtt{acred}^\mathtt{+}$ & \textbf{0.716} & 0.674 & 0.601 & \textbf{0.713} \\
    CNN & 0.629 & \textbf{0.807} & 0.456 & 0.583 \\
    SAF/S & 0.654 & 0.600 & 0.789 & 0.681 \\
    \hline
  \end{tabular}
\end{wraptable}

On FakeNewsNet, $\mathtt{acred}^\mathtt{+}$ obtained state of the art results and \texttt{acred} obtained competitive results in line with strong baseline systems reported in the original paper~\cite{shu2018FakeNewsNet}, shown in table~\ref{tab:fakeNewsNetResults}. We only consider as baselines systems which only use the article content, since acred does not use credibility reviews based on social context yet. Note that baselines used 80\% of the data for training and 20\% for testing, while we used the full dataset for testing.

We performed a manual error analysis on the \texttt{acred} results for FakeNewsNet\footnote{As stated above, we used the results of this analysis to inform the changes implemented in $\mathtt{acred}^\mathtt{+}$}:
\begin{itemize}
\item 29 errors of fake news predicted as highly credible ($r \ge 0.5$): 16 cases (55\%) were due to acred finding pre-crawled sentence matches in that appeared in \texttt{snopes.com}, but not \texttt{ClaimReview}s for that article. A further 7 cases (24\%) were due to finding \texttt{unrelated} sentences and using the \texttt{WebSiteCR} where those sentences appeared, while being over-confident about those credibilities.
  
\item 41 errors of fake news predicted with low-confidence ($c <= 0.7$). 30 of these cases (73\%) are due to the FakeNewsNet crawler as it fails to retrieve valid content for the articles: GDPR or site-for-sale messages instead of the original article content. In these cases, acred is correct in having low-confidence credibility ratings. In the remaining 27\% of cases, acred indeed failed to find evidence to decide on whether the article was credible or not.

\item 264 real articles were rated as being highly not credible ($r < -0.5$). This is by far the largest source of errors. We manually analysed 26 of these, chosen at random. In 13 cases (50\%), the real stance should be \emph{unrelated}, but is predicted as \emph{discussed} or even \emph{agrees}; often the sentences are indeed about a closely related topics, but still about unrelated entities or events. In a further 7 cases (27\%) the stance is correctly predicted to be \emph{unrelated}, but the confidence still passess the threshold. Hence 77\% of these errors are due to incorrect or overconfident linking by the $\mathtt{LinkBot}^{\mathtt{SemSim}}_\mathtt{Sentence}$.

\item 48 real articles were rated as being somewhat not credible ($-0.25 < r \le 0.25$), in a majority of these cases, the $r$ was obtained from $\mathtt{LinkBot}^{\mathtt{PreCrawled}}_{\mathtt{Sentence}}$ rather than from $\mathtt{LinkBot}^{\mathtt{SemSim}}_\mathtt{Sentence}$
\end{itemize}

\begin{figure}
  \centering
  \begin{subfigure}[b]{0.3\textwidth}
    \centering
    \includegraphics[width=\textwidth]{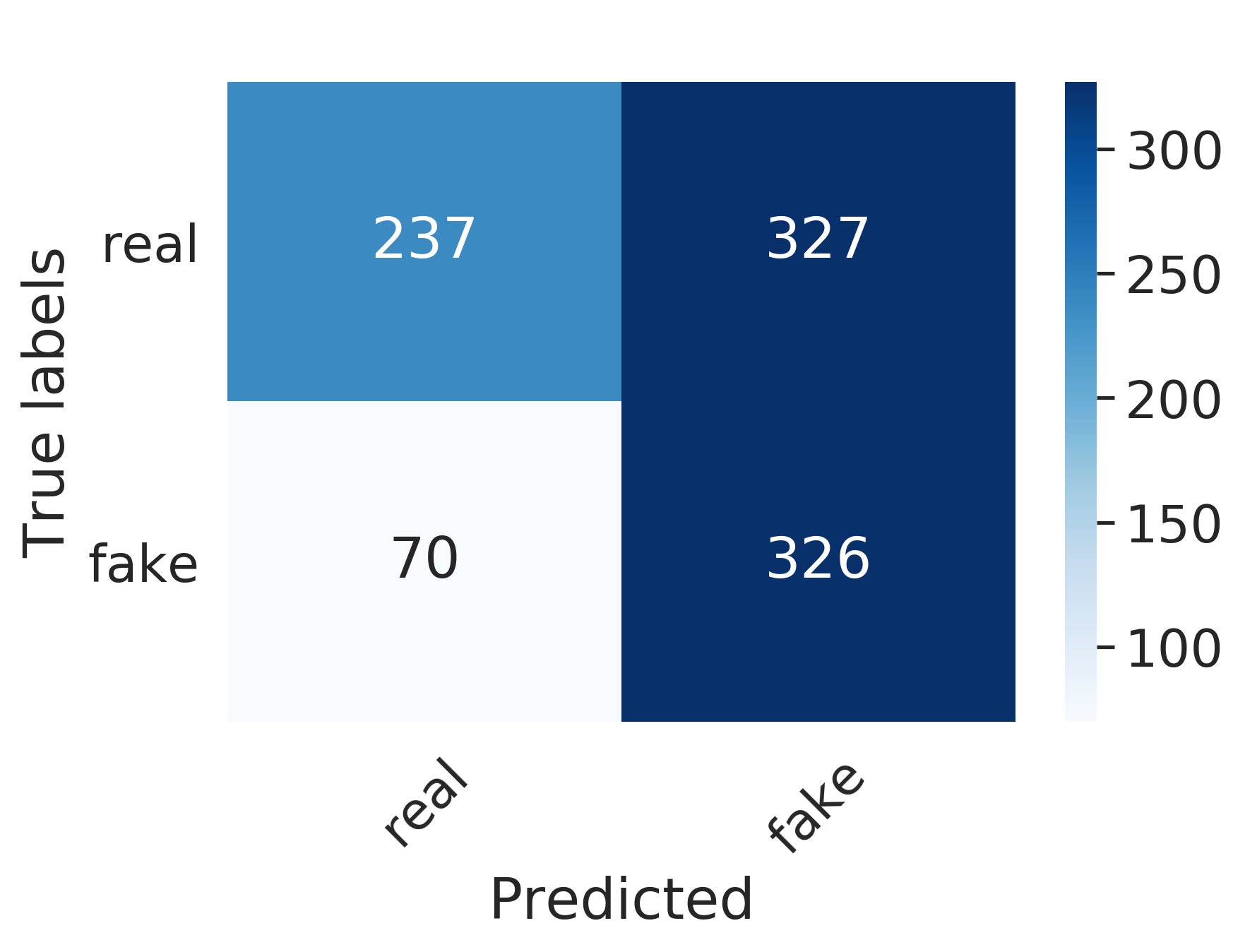}
    \caption{FakeNewsNet}
    \label{fig:cmFakeNewsNet}
  \end{subfigure}
  \hfill
  \begin{subfigure}[b]{0.3\textwidth}
    \centering
    \includegraphics[width=\textwidth]{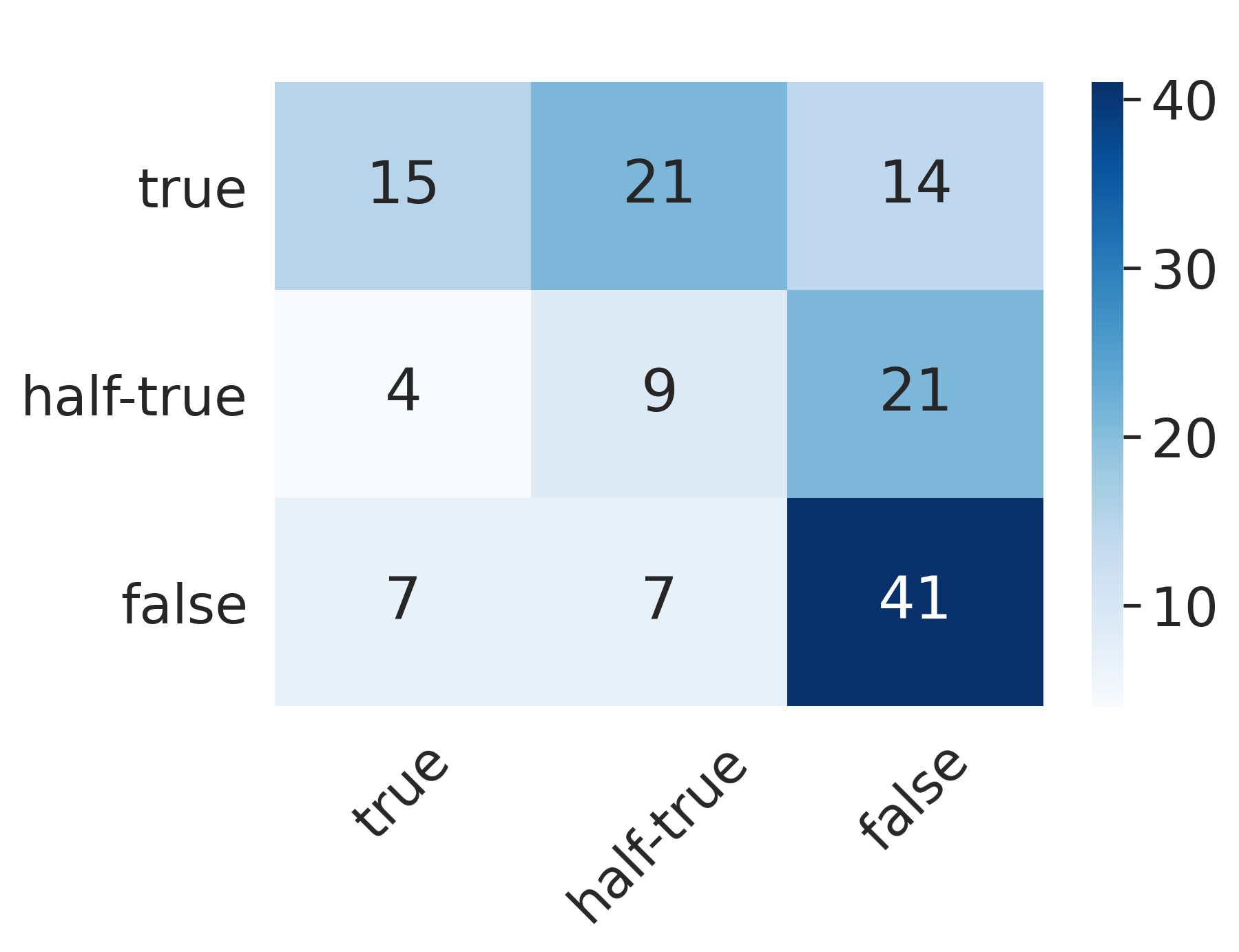}
    \caption{\texttt{clef18 (test)}}
    \label{fig:cmClef18}
  \end{subfigure}
  \hfill
  \begin{subfigure}[b]{0.3\textwidth}
    \centering
    \includegraphics[width=\textwidth]{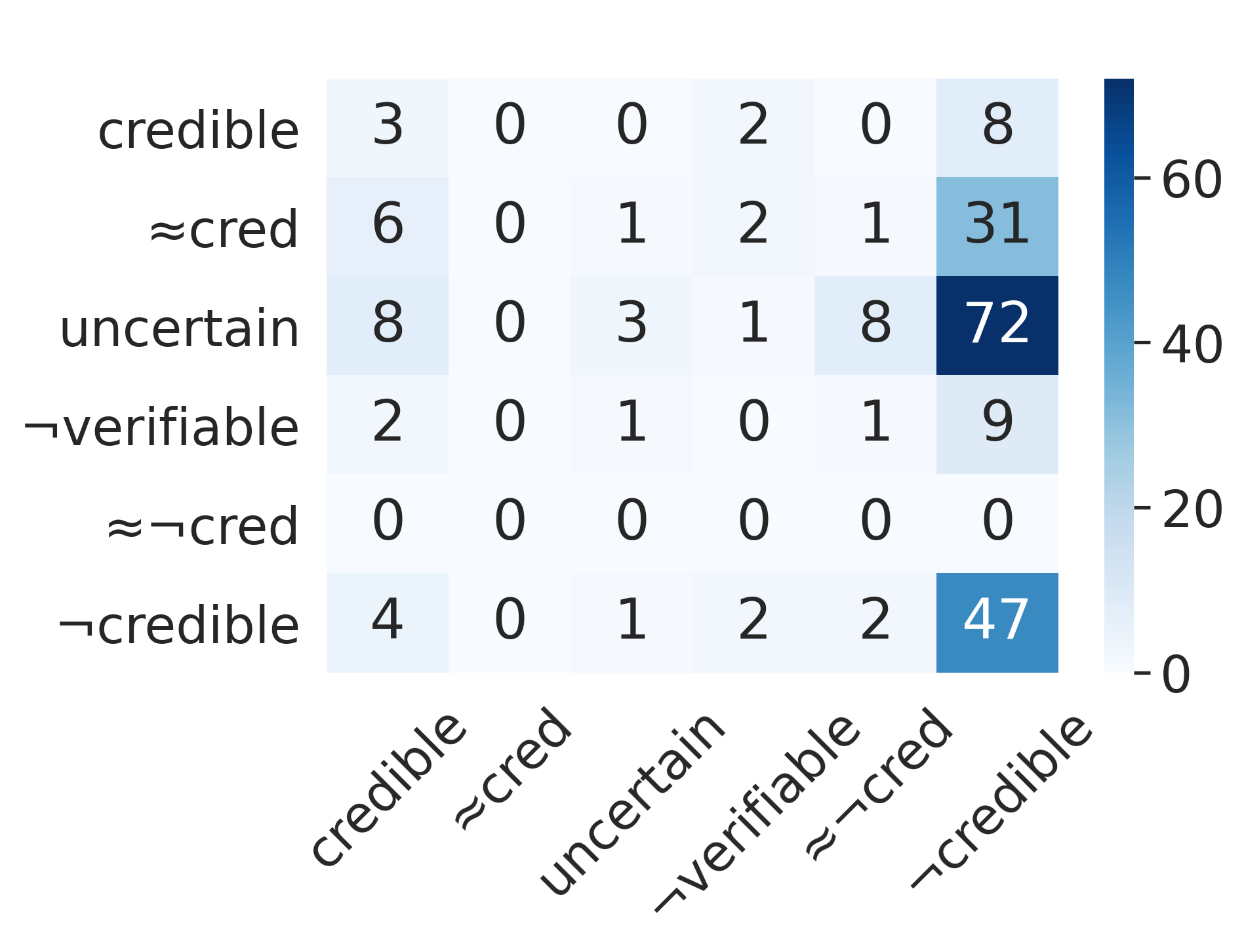}
    \caption{\texttt{coinform250}}
    \label{fig:cmTweets}
  \end{subfigure}
  \\
  \centering
  \begin{subfigure}[b]{0.3\textwidth}
    \centering
    \includegraphics[width=\textwidth]{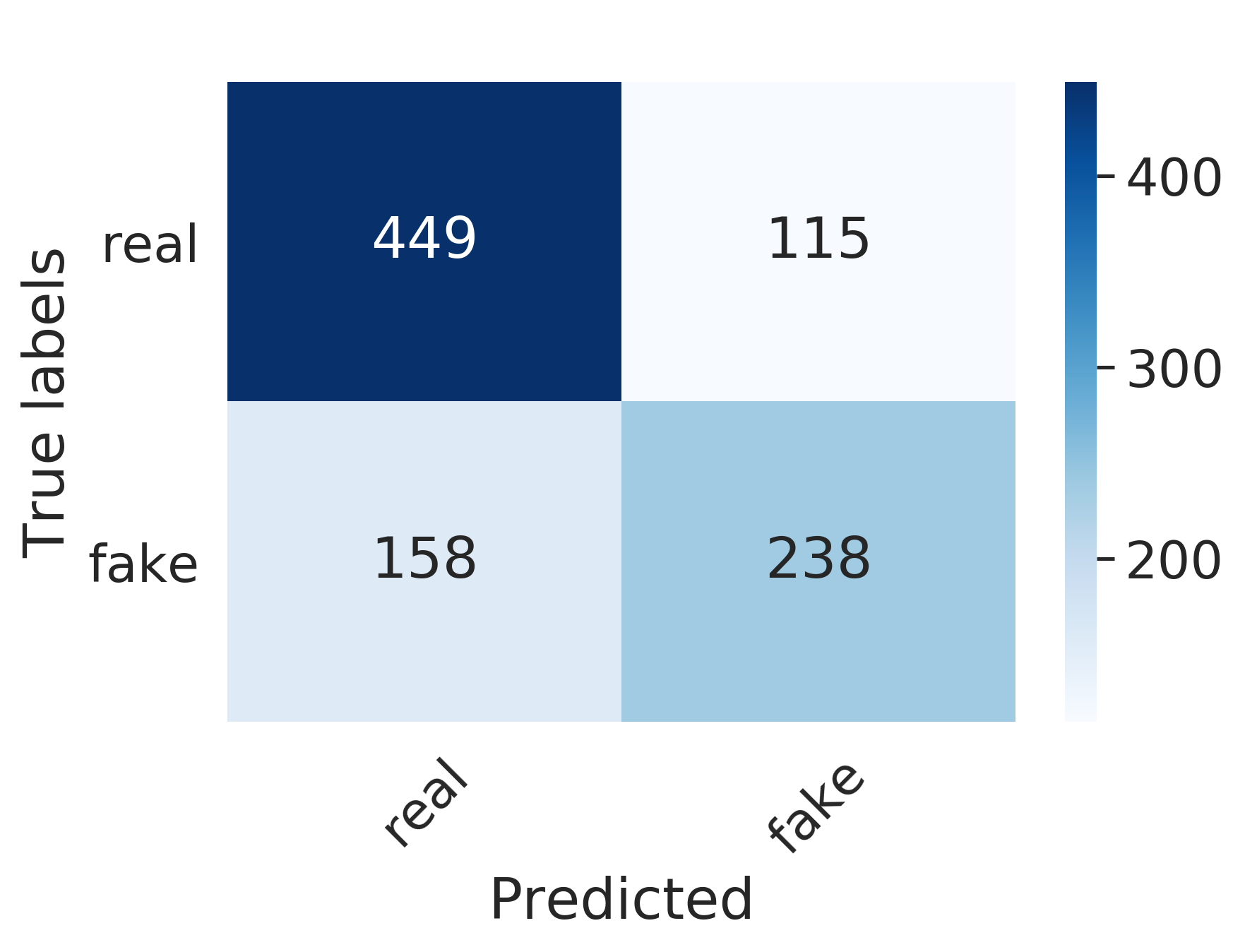}
    \caption{FakeNewsNet}
    \label{fig:cmFakeNewsNet_p}
  \end{subfigure}
  \hfill
  \begin{subfigure}[b]{0.3\textwidth}
    \centering
    \includegraphics[width=\textwidth]{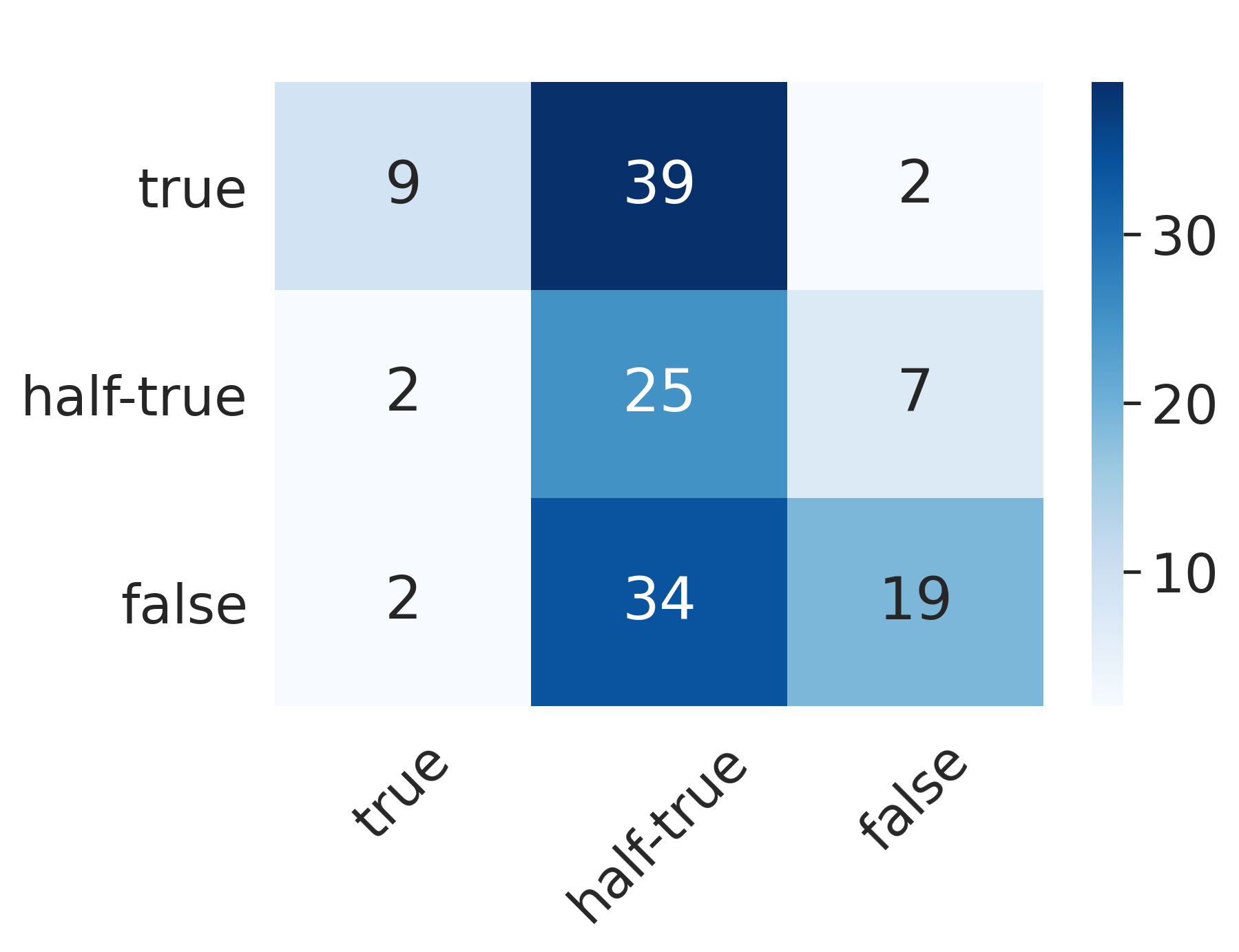}
    \caption{\texttt{clef18 (test)}}
    \label{fig:cmClef18_p}
  \end{subfigure}
  \hfill
  \begin{subfigure}[b]{0.3\textwidth}
    \centering
    \includegraphics[width=\textwidth]{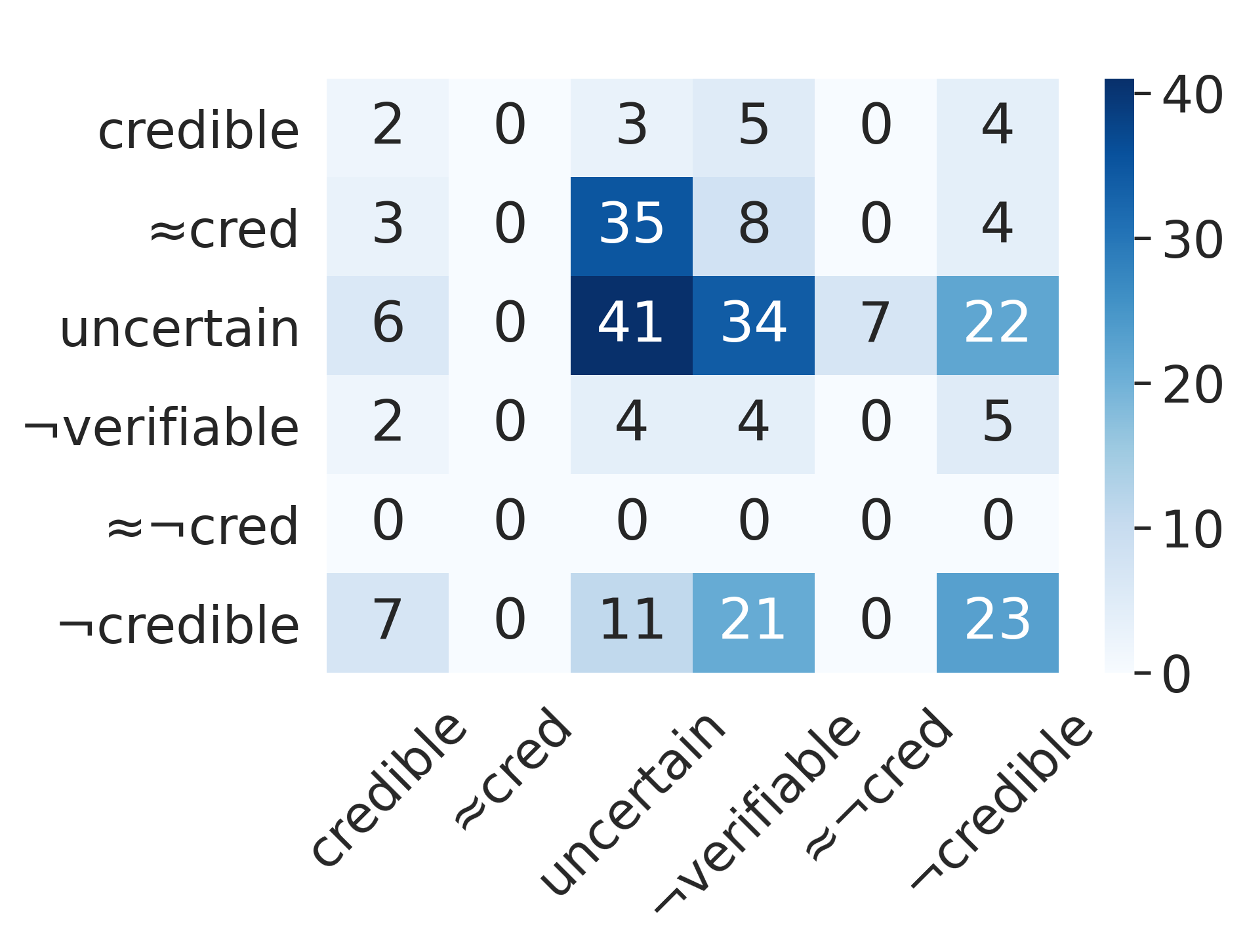}
    \caption{\texttt{coinform250}}
    \label{fig:cmTweets_p}
  \end{subfigure}
  \caption{Confusion matrices for \texttt{acred} (top) and $\mathtt{acred}^\mathtt{+}$ (bottom row) in evaluation datasets. We use $\approx$ for \emph{mostly} and $\neg$ for \emph{not} in the \texttt{coinform250} labels.}
  \label{fig:confmats}
\end{figure}

Finally, for the \texttt{coinform250} dataset, $\mathtt{acred}^\mathtt{+}$ obtains $0.279$ accuracy which is well above a baseline of random predictions, which obtains $0.167$ accuracy. The confusion matrix shown in figure~\ref{fig:cmTweets_p} shows that the performance is in line with that shown for FakeNewsNet (fig.\ref{fig:cmFakeNewsNet_p}) and \texttt{clef18} (fig.\ref{fig:cmClef18_p}). It also shows that \texttt{acred} tends to be overconfident in its predictions, while $\mathtt{acred}^\mathtt{+}$ is more cautious.

\subsection{Discussion}
\label{discussion}

\paragraph{The Good}
Our approach obtains competitive results in challenging datasets; these results are especially impressive when you take into account that we do not train or fine-tune our underlying models on these datasets. With $\mathtt{acred}^\mathtt{+}$, we also showed we can substantially improve results by performing some simple optimization of aggregation functions; however doing this in a systematic manner is not in the scope of this paper where we are focusing on validating the LCR design. Since the results were consistent across the different datasets, this shows that our network of bots have certain domain independence and validate our design for composable CR bots as our lookup and linking bots can successfully be reused by the high-level decomposing bots. We think this is largely due to our choice of generic deep-learning models for linguistic tasks like semantic similarity and stance detection where fairly large datasets are available.

We obtain state of the art results in all datasets. This shows that our approach excels at identifying misinformation, which is arguably the primary task of these kinds of systems. Furthermore, in many cases, our system correctly matches the misinforming sentence to a previously fact-checked claim (and article). Even when no exact match is found, often the misinforming sentence is linked to a similar claim that was previously reviewed and that is indicative of a recurring misinforming narrative. Such cases can still be useful for end-users to improve their awareness of such narratives.

\paragraph{The Bad}
The \texttt{acred} implementation is overly sceptical, which resulted in poor precision for data items which are (mostly) accurate or not verifiable. This is an important type of error preventing real-world use of this technology. As prevalent as misinformation is, it still only represents a fraction of the web content and we expected such errors to undermine confidence in automated systems. The presented error analysis shows that this is largely due to (i) errors in the stance-detection module or (ii) incorrect weighting of predicted stance and semantic similarity. The former was surprising as the stance prediction model obtained 92\% accuracy on FNC-1. This seems to be due to the fact that FNC-1 is highly unbalanced, hence errors in under-represented classes are not reflected in the overall accuracy. Note that we addressed this issue to a certain extent with $\mathtt{acred}^\mathtt{+}$, which essentially \texttt{compensates} for errors in the underlying semantic similarity and stance prediction modules.

The poor precision on real news is especially apparent in FakeNewsNet and \texttt{coinform250}. We think this is due to our naive implementation of our pre-crawled database of sentences extracted from websites. First, the relevant sentence extractor does not ensure that the sentence is a factual claim, therefore introducing significant noise. This suggests that the system can benefit from adding a check-worthiness filter to address this issue. The second source of noise is our selection of pre-crawled article sources which did not seem to provide relevant matches in most cases. We expect that a larger and more balanced database of pre-crawled sentences, coming from a wider range of sources, should provide a better pool of credibility signals and help to improve accuracy.

\paragraph{The Dubious}
Fact-checking is a recent phenomenon and publishing fact-checked claims as structured data even more so. It is also a laborious process that requires specialist skills. As a result, the pool of machine-readable high-quality reviewed data items is relatively small. Most of the datasets being used to build and evaluate automated misinformation detection systems are therefore ultimately based on this same pool of reviews. A percentage of our results may be based on the fact that we have exact matches in our database of ClaimReviews. On manual inspection, this does not appear to occur very often;  we estimate low, single digit, percentage of cases.

Although we did a systematic error analysis, presented in the previous section, we have not yet done a systematic success analysis. Cursory inspection of the successful cases shows that in some cases, we predicted the label correctly, but the explanation itself is incorrect; this tends to happen when a sentence is matched but is deemed to be \emph{unrelated} to a matched sentence. Note that other systems based on machine learning may suffer of the same issue, but unlike our approach they behave as black boxes making it difficult to determine whether the system correctly identified misinforming content for the wrong reasons.

\section{Conclusion and Future Work}
\label{conclusion}
In this paper we proposed a simple data model and architecture for using semantic technologies (linked data) to implement composable bots which build a graph of Credibility Reviews for web content. We showed that \texttt{schema.org} provides most of the building blocks for expressing the necessary linked data, with some crucial extensions. We implemented a basic fact-checking system for sentences, social media posts and long-form web articles using the proposed LCR architecture and validated the approach on various datasets. Despite not using the training sets of the datasets, our implementations obtained state of the art results on the \texttt{clef18} and FakeNewsNet datasets.

Our experiments have demonstrated the capabilities and added value of our approach such as human-readable explanations and machine aided navigation of provenance paths to aid in error analysis and pinpointing of sources of errors. We also identified promising areas for improvement and further research. We plan to (i) further improve our stance detection model by fine-tuning and testing on additional datasets\cite{Schiller2020}; (ii) perform an ablation test on an improved version of acred to understand the impact of including or omitting certain bots and datasets; (iii) perform crowdsourcing to evaluate both the understandability, usefulness and accuracy of the generated explanations. Beside our own plans, it is clear that a single organization or team cannot tackle all the issues which need to be resolved to achieve high-accuracy credibility analysis of web content. This is exactly why we propose the Linked Credibility Review, which should enable the distributed collaboration of fact-checkers, deep-learning service developers, database curators, journalists and citizens in building an ecosystem where more advanced, multi-perspective and accurate credibility analyses are possible.

\paragraph{Acknowledgements} 
Work supported by the European Comission under grant 770302 -- Co-Inform -- as part of the Horizon 2020 research and innovation programme. \includegraphics[width=0.05\textwidth]{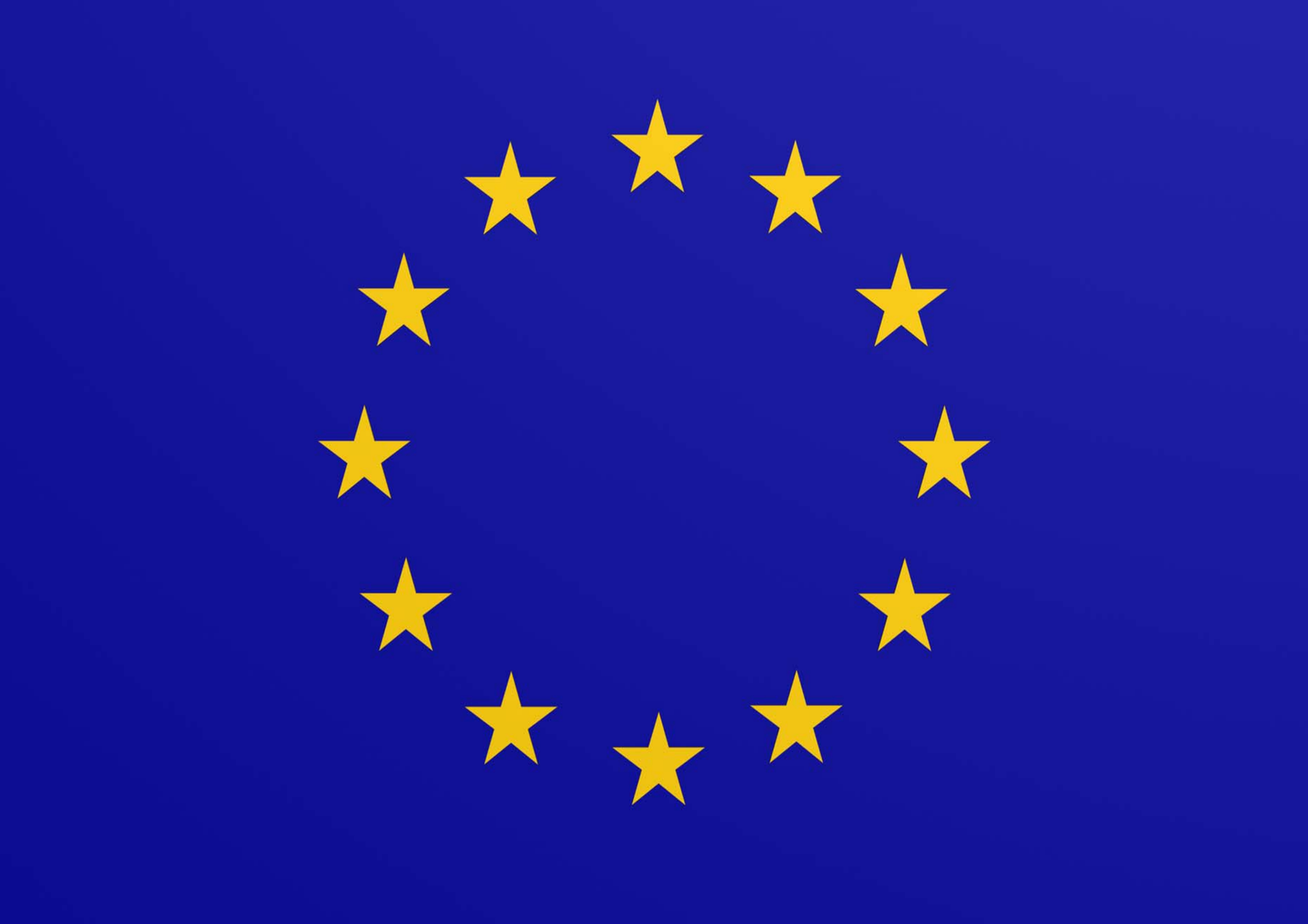} Thanks to Co-inform members for discussions which helped shape this research and in particular to Martino Mensio for his work on MisInfoMe. Also thanks to Flavio Merenda and Olga Salas for their help implementing parts of the pipeline. 

%
%
%
\bibliographystyle{splncs04}
\bibliography{acred_iswc20}

\end{document}